\documentclass[11pt,letterpaper,twoside,reqno,nosumlimits]{amsart}

\usepackage{etoolbox}

\patchcmd{\section}{\scshape}{\bfseries}{}{}
\makeatletter
\renewcommand{\@secnumfont}{\bfseries}
\makeatother

\usepackage[dvipsnames]{xcolor}
\patchcmd{\section}{\normalfont}{\normalfont\color{MidnightBlue}}{}{}
\patchcmd{\subsection}{\normalfont}{\normalfont\color{MidnightBlue}}{}{}
\patchcmd{\subsubsection}{\normalfont}{\normalfont\color{MidnightBlue}}{}{}

\makeatletter
\def\subsubsection{\@startsection{subsubsection}{3}%
\z@{.5\linespacing\@plus.7\linespacing}{-.5em}%
{\normalfont\bfseries}}
\makeatother

\usepackage{fancyhdr}
\usepackage{amsmath,amsfonts,amsbsy,amsgen,amscd,mathrsfs,amssymb,amsthm,mathtools,tensor}

\usepackage[a4paper]{geometry}
\usepackage{tikz}

\usepackage{stmaryrd}

\usepackage{caption}
\usepackage{graphicx}
\usepackage{color}
\usepackage[bf,SL,BF]{subfigure}
\usepackage{url}
\usepackage{epstopdf}
\usepackage{pdfsync}
\usepackage[colorlinks]{hyperref}
\usepackage[all]{xypic}
\usepackage{comment}
\usepackage{mathabx}
\usepackage{upgreek}

\usepackage{mathrsfs}
\usepackage{yfonts}

\hypersetup{
    linkcolor=blue,
}

\newlength{\fixboxwidth}
\newtheorem{example}{Example}

\usepackage{epsfig,amsbsy,graphicx,multirow}

\usepackage{ algorithm, algorithmic}
\renewcommand{\algorithmiccomment}[1]{\bgroup\hfill//~#1\egroup}

\usepackage[all,cmtip]{xy}

\usepackage{accents}

 \numberwithin{equation}{section}

\makeatletter

\usepackage{graphicx}

\makeatother

\usepackage{babel}

\usepackage{upgreek}

\begin{document}
\title{Decision theoretic bootstrapping}
\author[Tavallali, Bajgiran, Esaid, Owhadi]{Peyman Tavallali$^{1,*}$, Hamed Hamze Bajgiran$^{2}$, Danial J. Esaid$^{3}$, Houman Owhadi$^{2,*}$}
\begin{abstract}
The design and testing of supervised machine learning models combine
two fundamental distributions: (1) the training data distribution
(2) the testing data distribution. Although these two distributions
are identical and identifiable when the data set is infinite; they
are imperfectly known (and possibly distinct) when the data is finite
(and possibly corrupted) and this uncertainty must be taken into account
for robust Uncertainty Quantification (UQ). We present a general decision-theoretic bootstrapping solution to this problem:
(1) partition the available data into a training subset and a UQ subset
(2) take $m$ subsampled subsets of the training set and train $m$ models
(3) partition the UQ set into $n$ sorted subsets and take a random fraction 
of them to define $n$ corresponding empirical distributions $\mu_{j}$ 
(4) consider the adversarial game
where Player I selects a model $i\in\left\{ 1,\ldots,m\right\} $,
Player II selects the UQ distribution $\mu_{j}$ and Player I receives
a loss defined by evaluating the model $i$ against data points sampled
from $\mu_{j}$ (5) identify optimal mixed strategies (probability
distributions over models and UQ distributions) for both players.
These randomized optimal mixed strategies provide optimal model mixtures and
UQ estimates given the adversarial uncertainty of the training and
testing distributions represented by the game. The proposed approach provides (1) some degree of robustness to distributional shift in both the distribution of training data and that of the testing data (2) conditional probability distributions on the output space forming aleatory representations of the uncertainty on the output as a function of the input variable.
\end{abstract}

\maketitle
$^{1}$Jet Propulsion Laboratory, California Institute of Technology,
Pasadena, CA 91109

$^{2}$California Institute of Technology, Pasadena, CA 91125

$^{3}$Jet Propulsion Laboratory, California Institute of Technology,
Pasadena, CA 91109; University of California, Riverside, 92521

$^{*}$Corresponding Author, E-mails: peyman.tavallali@jpl.caltech.edu;
owhadi@caltech.edu

\textsuperscript{\textcopyright} 2021. California Institute of Technology. Government sponsorship
acknowledged. 

\section{Introduction\label{sec:Introduction}}

Although traditional cross validation (CV)  provides model UQ/error estimates \cite{stone1974cross,stone1977asymptotic,allen1974relationship,golub1979generalized,wahba1980spline}, these estimates are in general not  robust to distribution shifts. We propose to incorporate 
the impact of distributional uncertainty in the adversarial setting of decision theory (DT) \cite{owhadi2015towards}.

\begin{figure}[t]
\includegraphics[width=0.3\textwidth]{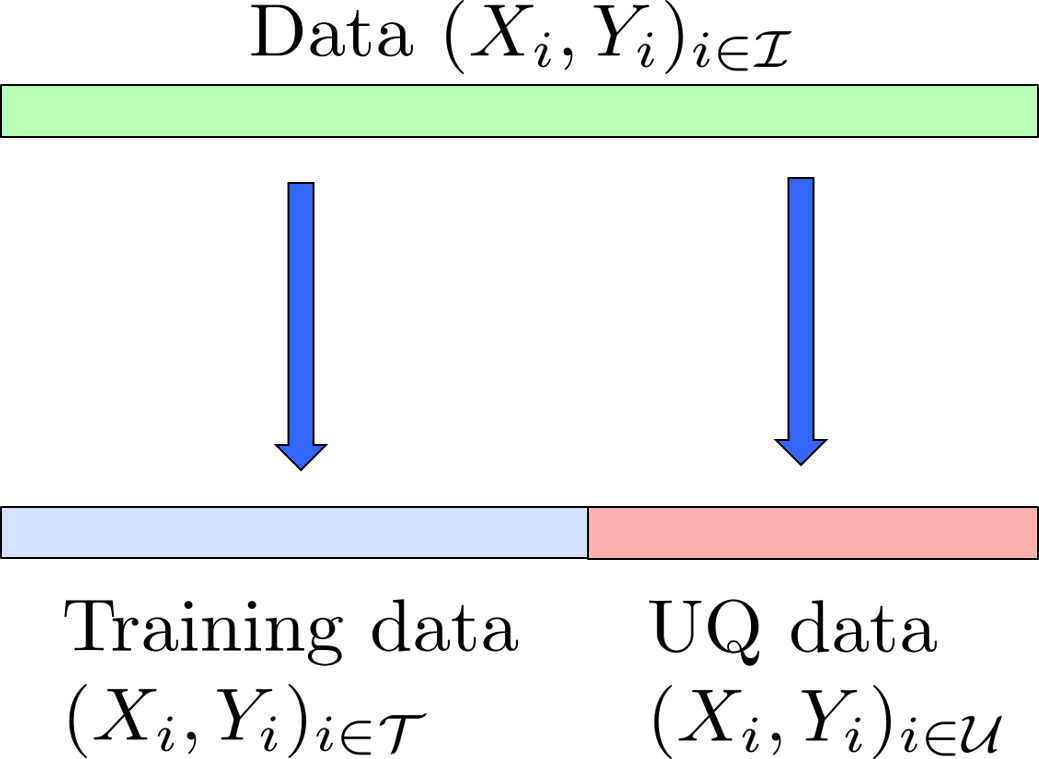}\,\,\,\includegraphics[width=0.3\textwidth]{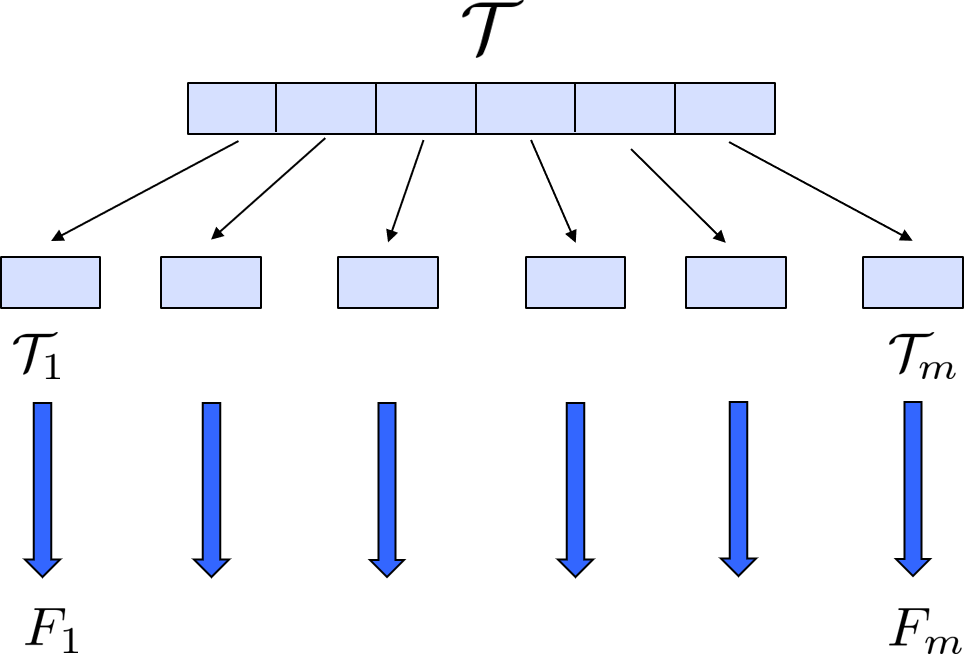}\,\,\,
\includegraphics[width=0.3\textwidth]{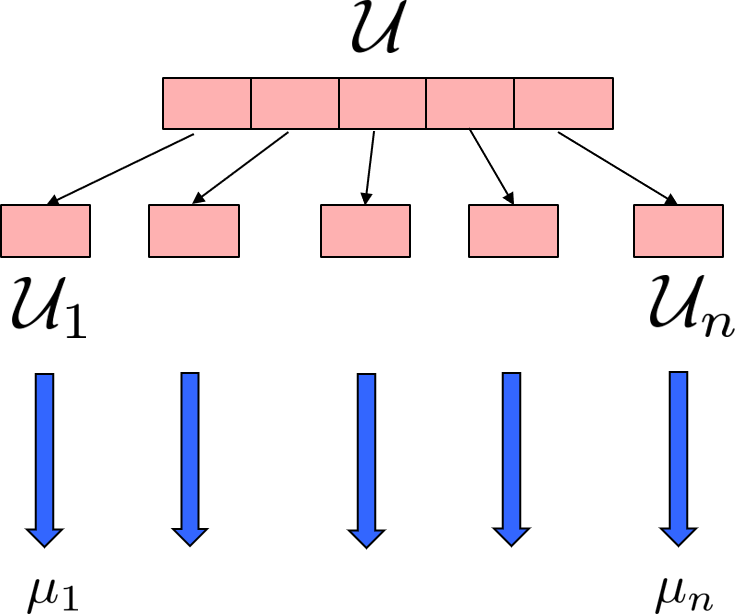}

\caption{\textbf{Train-UQ Partitioning of the Data}. The \emph{left plot} shows
the schematic partitioning of the original empirical data into two
exclusive train and UQ sets. The \emph{middle plot} shows how the
train set is used to train $m$ ML estimators (randomly). Here, the
training sets are found by subsampling the train dataset $m$ times. 
The \emph{right plot} shows the procedure of breaking
up the UQ set into $n$ subsets, from which $n$ different empirical
distributions can be extracted.}

\label{Fig:Train_UQ_Part}
\end{figure}

\subsection{Main Result:\label{subsec:Main-Result}}
Our proposed decision theoretic bootstrapping  (DTB)
solution  is to: 
\begin{itemize}
\item randomly partition the available data $\left\{ \left(X_{i},Y_{i}\right)\right\} _{i\in\mathcal{I}}$
into a training subset $\left\{ \left(X_{i},Y_{i}\right)\right\}_{i\in\mathcal{T}}$
and a UQ subset $\left\{ \left(X_{i},Y_{i}\right)\right\}_{i\in\mathcal{U}}$
(see Figure \ref{Fig:Train_UQ_Part}-a);
\item randomly subsample  the training set $\mathcal{T}$ into $m$ subsets $\left\{ \mathcal{T}_{i}\right\} _{i=1}^{m}$,
and train $m$ models $\left\{ F_{i}\right\} _{i=1}^{m}$ (see Figure
\ref{Fig:Train_UQ_Part}-b);
\item For $k\in \{1,\ldots,K\}$ repeat the following steps.
\item Partition the UQ set into $n$ sorted subsets.
\begin{itemize}
\item Randomly subsample from all the $n$ sorted subsets of $\mathcal{U}$
and define $n$ corresponding empirical distributions $\left\{ \mu_{j}\right\} _{j=1}^{n}$
(see Figure \ref{Fig:Train_UQ_Part}-b);
\item consider the zero-sum adversarial game where Player I selects a model
$F_{i}$, Player II selects the UQ distribution $\mu_{j}$, and Player
I receives a random loss defined as the empirical risk
\begin{equation}
\mathcal{L}_{ij}=\mathbb{E}_{(X,Y)\sim \mu_j}\mathcal{E}\left(Y,F_{i}\left(X\right)\right),\label{eq: Game Loss}
\end{equation}
where  $\mathcal{E}$ is an error function.  The diagram of the proposed game is as follows.
\[
\text{\ensuremath{\xymatrix{ \text{(Player I)}  &  F_{i}\ar[dr]_{\min}  &   &  \mu_{j}\ar[ld]^{\max}  &  \text{(Player II)}\\
  &   &  \mathcal{L}_{ij}  &   &  
}
} }\,
\]
\item identify and record the optimal mixed strategies (probability distributions) $\boldsymbol{p}^{(k)}=\left(p_{1}^{(k)},\ldots,p_{m}^{(k)}\right)$
over models $\left\{ F_{i}\right\} _{i=1}^{m}$ and $\boldsymbol{q}^{(k)}=\left(q_{1}^{(k)},\ldots,q_{n}^{(k)}\right)$
over UQ distributions $\left\{ \mu_{j}\right\} _{j=1}^{n}$, for both
players by solving the minimax problem 
\begin{equation}
\begin{array}{c}
\underset{\boldsymbol{p}}{\min}\,\underset{\boldsymbol{q}}{\max}\sum_{i,j}p_{i}\mathcal{L}_{ij}q_{j}\end{array}.\label{eq: minimax - compact version}
\end{equation}
and record the value $v^{(k)}$ of the game \eqref{eq: minimax - compact version}.
\end{itemize}
\item Use  
the distribution of the $p^{(k)}$ and possibly $v^{(k)}$  for UQ analysis. 
\end{itemize}

\subsection{Structure of the paper:\label{subsec:Paper-Structure}}
The proposed methodology and algorithm is described in Section
 \ref{sec:Methodology-and-Algorithm}. In Section \ref{sec:Relevance-to-CV},
we discuss the relevance of DTB to CV. We then
present two detailed examples in Section \ref{sec:Examples}. We cover
related works in \ref{sec:Related-Works} and compare the proposed method with other  UQ for ML methods in the literature. 

\section{DTB Methodology and Algorithm\label{sec:Methodology-and-Algorithm}}

\begin{figure}[t]
\includegraphics[scale=0.5]{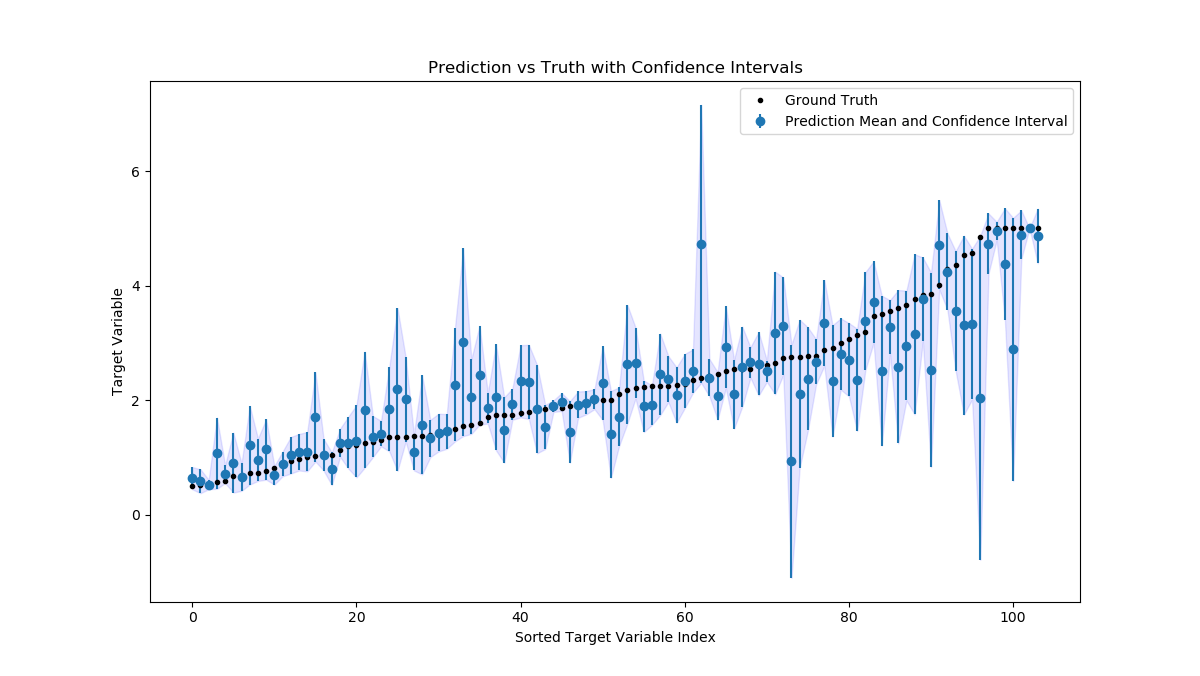}

\caption{\textbf{Robust Prediction with Confidence Intervals}. This plot shows
that on a random test set from the California Housing dataset \cite{pace1997sparse},
the predictions of $20$ different regression trees are robust in
the sense that their confidence intervals include the ground truth.
There are only a few large confidence intervals, and the rest are relatively
tight. This plot is showing the value of double randomization, both
over the empirical distributions and the models. The latter is one of the main ideas of the adversarial game-theoretic setup of DTB.}

\label{Fig: Pred_UQ}
\end{figure}

In any modeling endeavor, there are three major questions: ``\emph{How
can the uncertainty be quantified?}'', ``\emph{How robust is a given model if the distribution of the testing data is distinct from that of training data?}'', and finally,
``\emph{How can we pick amongst many trained models?}''. This manuscript presents a decision-theoretic
(zero-sum game) framework answering all three questions. We will now motivate these questions and illustrate DTB with the following real-world problem.
\begin{example}
Given the California housing dataset (that includes ZIP codes, square footage, number of bedroom and bathrooms, sales prices, etc.) \cite{pace1997sparse},
and $20$  ML models (e.g., regression
trees) trained on distinct subsets of this dataset,
(1) predict sales prices for a new test set (that includes the previous features but not sales prices), (2) provide confidence intervals for each prediction, (3) establish the robustness of the predictions to distributional shift between training and testing data.
Although selecting the model with the lowest error in the training set is a natural solution to (1), the deterministic nature of this solution is in conflict with the requirements of (2) providing confidence intervals and (3) ensuring robustness to distributional shifts.
Similarly, although using training errors to create a weighted ranking of the $20$ models could be used to
provide confidence intervals, it does not ensure robustness to distribution shifts.
With the proposed approach, we
estimate the error of one of these $20$ models (selected at random by Player I) against an
empirical distribution  (selected at random by player II). The randomization of the underlying game provides confidence intervals, and its adversarial nature ensures robustness to distribution shifts.
Figure \ref{Fig: Pred_UQ} illustrates DTB. For that figure, 
we trained $m=20$ different regression tree models (each with a maximum depth of $15$) on a train set with $|\mathcal{T}|=13759$ samples, then exposed those models
to $n=100$ unseen randomized realizations of a UQ set (with $|\mathcal{U}|=6777$ samples) to find each model's
probability, and finally applied them to a left-out test set with $104$ samples. We chose a small test set to accentuate uncertainties for ease of presentation. Each model is trained on a random half of the training set ($|F_i|\approx |\mathcal{T}|/2$). The sorted unseen randomized realizations of the UQ set are divided into $n=100$ subsets. The combined support of the $\mu_j$ in each randomization encompasses roughly $20\%$ of the whole UQ set. The \emph{prediction distribution} $p_i$ over the $20$ models  is calculated based on material presented in Section \ref{subsec:Prediction-UQ:}. For each $104$ sample in the left-out test set, Figure \ref{Fig: Pred_UQ} shows the mean prediction (obtained by averaging the prediction of each model with respect to the distribution $p_i$) and the standard deviation of that prediction (with respect to the distribution $p_i$). 
As can
be seen from this figure, almost all ground truth values over the
test set fall in between the confidence bounds of the predictions (note that we used one standard deviation to define confidence intervals). 
\end{example}

\subsection{Setup\label{subsec:Setup}}
The supervised ML setting is that of approximating an unknown function $F^\dagger$ given $F^\dagger(X_i)=Y_i$ for all $i$ where $\left\{ \left(X_{i},Y_{i}\right)\right\} _{i\in\mathcal{I}}$
represents the available input/output training data. 
One major issue with solving this problem as a regression/interpolation (curve fitting) problem is that the interpolant does not carry an intrinsic measure of uncertainty  (even if it has been obtained as a conditional expectation by conditioning a Gaussian prior \cite{rasmussen2003gaussian}, then the conditional covariance of the posterior distribution depends on the covariance function of the Gaussian prior whose selection is to some degree arbitrary).
Furthermore, this approach is not 
stable to distribution shifts, especially in deep neural networks
case \cite{ovadia2019can}. 
A partial remedy is to use CV to ensure that 
the training and generalization errors are close \cite{stone1974cross,stone1977asymptotic,allen1974relationship,golub1979generalized,wahba1980spline}.
However, CV does not address the distribution shift problem and it does not
provide a pointwise (input dependent) prediction confidence intervals. Although there are methods that provide a stable linear regression mitigating the effects of distribution shifts by a relaxed minimax approach \cite{bertsimas2020stable}, yet these methods do not provide a pointwise prediction confidence interval. Contrary to these, our proposed solution, based on decision theory \cite{owhadi2015towards}, is to formulate the underlying regression problem (linear or non-linear) as a zero-sum adversarial game over the set of possible machine learning
(ML) models and empirical data distributions providing both robustness of estimation and prediction confidence intervals. To do so, the modeler
must provide a set of fitting functions $\left\{ F_{i}\left(X\right)\right\} _{i=1}^{m}$
and play an adversarial zero-sum game against some random empirical
distributions of the data. In what follows, we present a detailed description of each step.

\subsection{Partitioning the training data:\label{subsec:Train-and-UQ}}

Our first step is to
 randomly partition the available (training) data $\left\{ \left(X_{i},Y_{i}\right)\right\} _{i\in\mathcal{I}}$
into two mutually exclusive sets: a training set $\left\{ \left(X_{i},Y_{i}\right)\right\} _{i\in\mathcal{T}}$ and a UQ set  $\left\{ \left(X_{i},Y_{i}\right)\right\} _{i\in\mathcal{U}}$. 

\subsection{Training models:\label{subsec:Finding-Acceptable-Fitting}}
Next, we subsample $m$  subsets of the training set $\mathcal{T}$. We write
 $\left\{ \mathcal{T}_{i}\right\} _{i=1}^{m}$ for these subsets and train $m$
 corresponding models $\left\{ F_{i}\right\} _{i=1}^{m}$ ($F_i$ is the model obtained by using $\mathcal{T}_{i}$ as training data). The 
 training pipeline can be arbitrary (e.g., the corresponding models  could be  deep 
neural networks \cite{goodfellow2016deep}, random forests
\cite{breiman2001random,ho1995random}, or any other ML model). Therefore, there
 is no constraint in the selection of the model producing pipeline and any function
estimator, relevant to the problem at hand, and any training methodology
could be used. Even the class of these functions could be different across $i\in \{1,\ldots,m\}$.
For example, some could be linear regression models (for some labels $i$), some decision
trees \cite{breiman1984classification} (for the other labels) and some support vector machines
\cite{cortes1995support} (for the remaining labels). 

\subsection{Bootstrapping a set of empirical distributions:\label{subsec:Finding-a-Set}}
Next, we sort $\mathcal{U}$ with respect to the response variable $Y$ and then partition this sorted set into $n$ subsets $\left\{ \mathcal{U}_{j}\right\} _{j=1}^{n}$. Then, we randomly subsample the sorted set $\left\{ \left(X_{i},Y_{i}\right)\right\} _{i\in\mathcal{U}}$ from the $n$ subsets $\left\{ \mathcal{U}_{j}\right\} _{j=1}^{n}$
and define $n$ corresponding empirical distributions 
\begin{equation}
\mu_{j}\coloneqq\frac{1}{\left|\mathcal{U}_{j}\right|}\sum_{j\in\mathcal{U}_{j}}\updelta\left(X_{j},Y_{j}\right).\label{eq:mu_j}
\end{equation}

\subsubsection{Deterministic Game:\label{subsec:Deterministic-Game-Situation:}}
Consider the game in which Player I  selects an element from $\left\{ F_{i}\right\} _{i=1}^{m}$
and Player II selects an element from $\left\{ \mu_{j}\right\} _{j=1}^{n}$ (defined as in \eqref{eq:mu_j}). 
Let  
\begin{equation}\label{eq: loss matrix}
\mathcal{L}_{ij}=\mathbb{E}_{(X,Y)\sim \mu_j}\big[\mathcal{E}(Y,F_i(X))\big]
\end{equation}
be the loss function for this game.
In that deterministic setting the optimal mixed strategies of Player I and II tends to concentrate on single elements, i.e., the worst distribution in $\left\{ \mu_{j}\right\}_{j=1}^{n}$ and its best response in   $\left\{ F_{i}\right\} _{i=1}^{m}$.
This is due to the fact that, although this game incorporates some degree of uncertainty in the selection of the data generating distribution, it does not incorporate the uncertainty associated with sampling from that distribution. 
To address this issue, we repeat the proposed game over many random realizations of the $\mu_j$ and use the resulting distribution over optimal mixed strategies for our UQ analysis.

\subsection{Randomized Game:\label{subsec:Setting-Up-the}}
Consider the game described in Subsection \ref{subsec:Deterministic-Game-Situation:} and 
observe that the randomized sampling process generating the $\mu_j$ induces a randomization of the loss metric $\mathcal{L}$ ($\mathcal{L}_{ij}$ a random function of $(i,j)$).
Assigning probabilities $\boldsymbol{p}=\left(p_{1},\ldots,p_{m}\right)$
over models $\left\{ F_{i}\right\} _{i=1}^{m}$ and $\boldsymbol{q}=\left(q_{1},\ldots,q_{n}\right)$
over a specific realization of $\left\{ \mu_{j}\right\} _{j=1}^{n}$,
 optimal mixed strategies $(p^\dagger,q^\dagger)$ are identified as solutions of the minimax problem
\begin{equation}
\begin{array}{c}
\underset{\boldsymbol{p}}{\min}\,\underset{\boldsymbol{q}}{\max}\sum_{i,j}p_{i}\mathcal{L}_{ij}q_{j}\end{array},\label{eq: minimax}
\end{equation}
Note that
the  randomization of the loss $\mathcal{L}$ implies that  $(\boldsymbol{p}^\dagger,\boldsymbol{q}^\dagger)$ are random distributions. We will now sample from these distributions by solving \eqref{eq: minimax} many times (with different random realizations of $\mathcal{L}$).
 This approach is directly related to 
Harsanyi's purification over repeated games \cite{harsanyi1973games}. 

\subsection{Probability distributions over models:\label{subsec:Finding-Mixed-Model}}
Given $K\geq 1$, let $(\boldsymbol{p}^{(k)},\boldsymbol{q}^{(k)})$ be $K$ i.i.d. samples from $(\boldsymbol{p}^\dagger,\boldsymbol{q}^\dagger)$ and write $v^{(k)}$ for the corresponding random values of \eqref{eq: minimax}. Note that the $(\boldsymbol{p}^{(k)},\boldsymbol{q}^{(k)})$ are obtained by solving \eqref{eq: minimax} over $K$ independent realizations of the loss $\mathcal{L}$. 
We
record each optimal $\boldsymbol{p}^{\left(k\right)}$, each value $v^{(k)}$, and use the histogram defined by the $v^{(k)}$ and
the empirical average
\begin{equation}
\bar{\boldsymbol{p}}=\frac{1}{K}\sum_{k=1}^{K}\boldsymbol{p}^{\left(k\right)}\,,\label{eq:Mixed Model Dist}
\end{equation}
for UQ.
From an ML perspective, this approach is robust to distribution shifts. This procedure is presented
in Algorithm \ref{Alg: DTCV}. 

\begin{algorithm}[t]
\begin{enumerate}
\item Randomly separate $\left\{ \left(X_{i},Y_{i}\right)\right\} _{i\in\mathcal{I}}$
into mutually exclusive sets $\left\{ \left(X_{i},Y_{i}\right)\right\} _{i\in\mathcal{T}}$
and $\left\{ \left(X_{i},Y_{i}\right)\right\} _{i\in\mathcal{U}}$.
\item Generate $m$ randomly subsampled subsets  $\left\{ \mathcal{T}_{i}\right\} _{i=1}^{m}$ of the training set $\mathcal{T}$. 
\begin{enumerate} 
\item Train the models $F_{i}$ on the subsets $\mathcal{T}_{i}$.
\end{enumerate}
\item Sort $\mathcal{U}$ with respect to the response variable $Y$ and then partition into $n$ subsets $\left\{ \mathcal{U}_{j}\right\} _{j=1}^{n}$.
\item For $k\in\left\{ 1,\ldots,K\right\} $
\begin{enumerate}
\item Generate $n$ randomly subsampled subsets $\left\{ \mathcal{U}_{j}^{(k)}\right\} _{j=1}^{n}$ from $\left\{ \mathcal{U}_{j}\right\} _{j=1}^{n}$ write
$\left\{ \mathcal\mu_{j}^{(k)}\right\} _{j=1}^{n}$ for the corresponding empirical distributions.
\item Solve the minimax problem $\begin{array}{c}
\underset{\boldsymbol{p}}{\min}\,\underset{\boldsymbol{q}}{\max}\sum_{i,j}p_{i}\mathcal{L}_{ij}q_{j}\end{array}$, and record the optimal $\boldsymbol{p}^{\left(k\right)}$ and the value $v^{(k)}$ of the game.
\end{enumerate}
\item Report $\bar{\boldsymbol{p}}=\frac{1}{K}\sum_{k=1}^{K}\boldsymbol{p}^{\left(k\right)}$
as the  distribution over models and the histogram of values $v^{(k)}$.
\end{enumerate}
\caption{\textbf{Decision Theoretic CV}}
\label{Alg: DTCV}
\end{algorithm}

\subsection{How to Choose $K$, $n$ and the $\mathcal{U}_j$:\label{subsec:How-to-Choose}}

Choosing $K$ is relatively straightforward. Since (\ref{eq:Mixed Model Dist}) is a Monte Carlo (MC) estimate, its accuracy improves with the number of sample points. The $\mathcal{U}_j$ are obtained through random subs-sampling of $\mathcal{U}$ and are of chosen to be of equal size $s$ ($|\mathcal{U}_j|=s$). 
The $\mathcal{U}_j$ do not overlap. We selected $s$ and $n$ so that the size of the combined support 
of the $\mathcal{U}_j$ is about $20\%$ of the size of $\mathcal{U}$  ($n s \approx 0.2 |\mathcal{U}|$). Note that given 
$n s \approx 0.2 |\mathcal{U}|$, the  parameterization of our 
approach can be reduced to $s$ which can be interpreted as a trade-off parameter between robustness (small $s$) and accuracy (large $s$).
Recall that robustness and accuracy are conflicting requirements \cite{owhadi2017qualitative}.

\section{UQ analysis\label{sec:Relevance-to-CV}}
\subsection{CV and Bayesian inference}
While CV provides an estimate of the generalization error, this estimate is not pointwise in the sense that it does not depend on the particular input of the model. Furthermore, CV does not produce any distribution over models. Furthermore, while Bayesian inference could be employed to obtain a distribution over models, doing so would not be robust with respect to the selection of the prior  \cite{owhadi2015brittleness,owhadi2015brittlenessb,owhadi2016selberg, owhadi2017qualitative}.

\subsection{DTB\label{subsec:Prediction-UQ:}}
On the other hand, by inducing a probability distribution $\bar{p}$ over models, DTB defines a robust point-wise distribution over the output space where robustness in inherited from the adversarial nature of the game. This distribution depends on the particular choice of the loss function  and  the histogram of the 
 $\left\{ v^{\left(k\right)}\right\} _{k=1}^{K}$ provides a robust quantification of generalization error.
Given an input $X$, write 
\begin{equation}
\rho\left(X\right)=\sum_{i=1}^{m}F_{i}\left(X\right)\bar{p}_{i}\,,\label{eq: empirical expected}
\end{equation}
 for the average model output and 
\begin{equation}
\sigma^{2}\left(X\right)=\sum_{i=1}^{m}\left(F_{i}\left(X\right)-\rho\left(X\right)\right)^{2}\bar{p}_{i}\,,\label{eq: empirical std}
\end{equation}
for the standard deviation of the model outputs defined by  $\bar{\boldsymbol{p}}$. Note that $\bar{\boldsymbol{p}}$ may be distinct from the uniform distribution employed in \emph{ensemble modeling}.

\section{Examples\label{sec:Examples}}
\begin{figure}[t]
\includegraphics[scale=0.3]{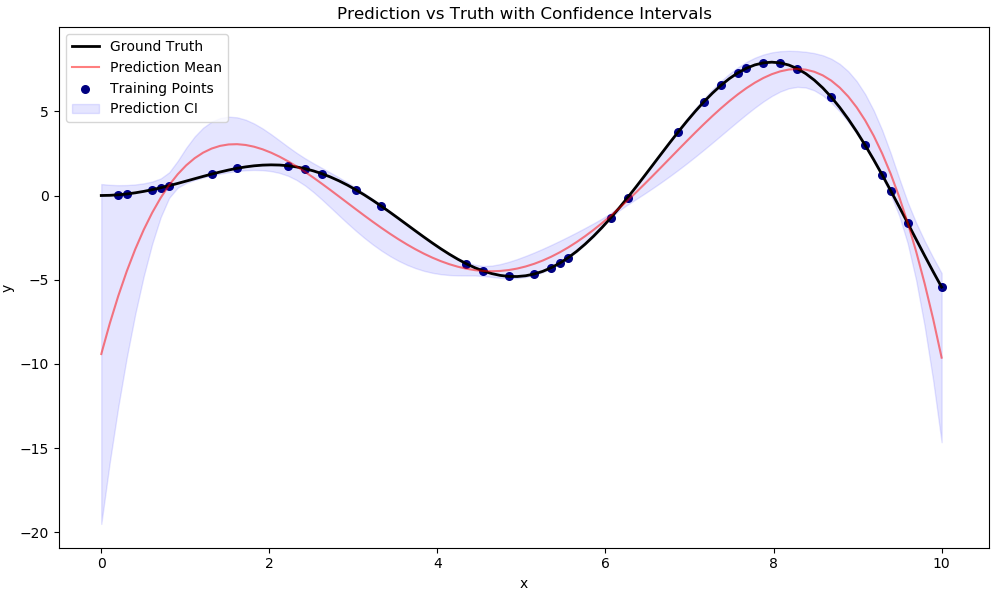}\includegraphics[scale=0.3]{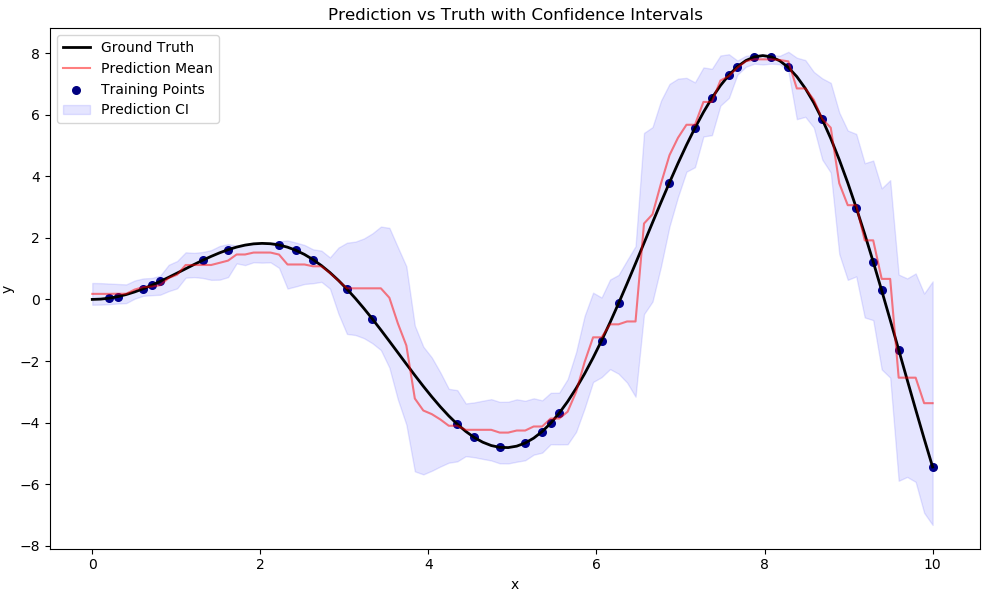}

\caption{\textbf{Robust Prediction with Confidence Intervals}. The \emph{left
plot} shows the application of Algorithm \ref{Alg: DTCV} when the
ML estimators are $20$ regression polynomials each with degree $4$.
The ground truth is shown in solid dark color. The sample points are
the dots, and the prediction mean and confidence intervals are shown
in red and filled-blue, respectively. The \emph{right plot} uses the
setting of the left plot, except that the $20$ regressors
are regression trees. Irrespective of the type of the regressors,
both plots show robust predictions and confidence intervals obtained from only a sparse set of non-uniform samples.}

\label{Fig: Pred_UQ_xsinx}
\end{figure}

We will now illustrate the proposed approach (Algorithm \ref{Alg: DTCV}) on synthetic and real-world datasets.
\begin{example}
Consider the univariate function $y=x\sin x$ for $x\in\left[0,10\right]$ (solid dark plots in Figure \ref{Fig: Pred_UQ_xsinx}). The dataset is composed of
 $35$ pointwise evaluations of this function. In the left plot of Figure \ref{Fig: Pred_UQ_xsinx} we regress each training subset $\mathcal{T}_i$ with a polynomial of degree $4$ to produce the corresponding model $F_i$ and use  Algorithm \ref{Alg: DTCV}
with $20$ regression polynomials to obtain a probability distribution $\bar{\boldsymbol{p}}$ over these models.
 As seen in this plot, the ground truth is close to the prediction mean, defined
by (\ref{eq: empirical expected}), and it falls between the confidence
bounds, defined by (\ref{eq: empirical std}). Although low order polynomials do not lead to highly accurate models, Algorithm
\ref{Alg: DTCV} provides a robust quantification of generalization errors. Replacing
the $20$ regressions polynomials by regression trees further improves this picture, as illustrated in the right plot of Figure \ref{Fig: Pred_UQ_xsinx}. In
this case, we fixed the depth of each tree to $5$. Note that the confidence intervals are now smaller near the boundaries.
Generally, this simple example illustrates the robustness of Algorithm \ref{Alg: DTCV}
 when the data is sparse,  non-uniformly distributed, and the models are weak. 
\end{example}

\begin{example}
With this example, illustrate the robustness of Algorithm \ref{Alg: DTCV} on the following datasets: California Housing \cite{pace1997sparse},
Electrical Grid Stability \cite{arzamasov2018towards}, Superconductor
\cite{hamidieh2018data}, and Bike Sharing \cite{fanaee2014event}.
For the sake of brevity, we call them Housing, Grid, SC, and Bike,
respectively. To normalize target ranges, the target variables in Grid, SC and
Bike were multiplied by $100.0$, $0.1$, and $0.01$, respectively.

For contrast, we compare the model distribution obtained from Algorithm \ref{Alg: DTCV} (DT models)  with the uniform distribution. In all
the cases, we take $10$ regression trees with a maximum depth of
$10$. The original data is divided into $80\%$ train-UQ and $20\%$
test sets. The train-UQ set is then fed into Algorithm \ref{Alg: DTCV}
once having separate $\mathcal{T}$ and $\mathcal{U}$ sets, and once
$\mathcal{T}=\mathcal{U}$. The latter is done for the sake of a fair
comparison with uniform ensembles. In the algorithm, we take $n=100$ and
$K=100$. Also, for each $k\in\left\{ 1,\ldots,K\right\} $,
we take $\frac{s n}{\left|\mathcal{U}\right|}\simeq0.2$. 

We introduce two modes of training: In the first mode, each tree is
trained on randomly selected $0.5\%$ of the train set $\mathcal{T}$.
In the second mode, each tree is trained on randomly selected $50\%$
of the train set. The first and second modes are called ``weak''
and ``strong'', respectively. 

Furthermore, we introduce two comparison metrics: One is the overall
mean squared loss on the test set. The other is the maximum mean squared
loss over folds, where folds are determined by binning the sorted
target variable. In these experiments, we use $100$ folds. In fact, this
second metric is an indicator of the robustness of the final ensemble
with respect to distribution shifts. 

We repeat each setup in this experiment $20$ times. The results are
reported in Tables \ref{Tab: Overall} and \ref{Tab: MaxFold}. The
values in parenthesis are the ones produced with $\mathcal{T}=\mathcal{U}$.
As seen from Table \ref{Tab: MaxFold}, the ensemble of estimators
generated by Algorithm \ref{Alg: DTCV} are showing lower maximum
fold losses in nearly all weak cases and most of the strong cases, compared to their uniform averaging
counterparts. The difference is more prominent when the models are
weakly trained. Having a lower maximum fold loss is an indication
of the robustness of Algorithm \ref{Alg: DTCV} with respect to distribution
shifts, especially when the models are not strong. In other words,
Algorithm \ref{Alg: DTCV} shows its best applicability when the epistemic
uncertainties are notable. On the other hand, most of the times, the
overall loss is lower for uniform averaging of the models; see Table
\ref{Tab: Overall}.  

It is noteworthy to mention that the improvement in the maximum fold
loss is larger than the degradation in the overall loss when Algorithm
\ref{Alg: DTCV} is used. In other words, if Algorithm \ref{Alg: DTCV}
is used, with a slight sacrifice of the overall accuracy, the models
become more stable and consistent, with higher precision, in their
predictions. 
\end{example}

\begin{table}[t]
\begin{tabular}{|c|c|c|c|c|}
\hline 
Overall & \multicolumn{1}{c}{Weak Trees} &  & \multicolumn{1}{c}{Strong Trees} & \tabularnewline
\hline 
 & Uniform & DT & Uniform & DT\tabularnewline
\hline 
Housing & \textbf{0.54}(\textbf{0.51}) & 0.59(0.66) & \textbf{0.31}(0.31) & 0.32(0.31)\tabularnewline
\hline 
Grid & 7.54(\textbf{7.23}) & \textbf{7.27}(7.58) & \textbf{1.95}(\textbf{1.92}) & 1.98(1.95)\tabularnewline
\hline 
SC & \textbf{3.41}(\textbf{3.15}) & 3.52(3.81) & \textbf{1.28}(1.23) & 1.30(\textbf{1.22})\tabularnewline
\hline 
Bike & \textbf{1.47}(\textbf{1.39}) & 1.66(1.87) & \textbf{0.28}(\textbf{0.27}) & \textbf{0.28}(0.28)\tabularnewline
\hline 
\end{tabular}

\caption{\textbf{Overall Losses Comparison}. This table shows the overall loss
comparison of 10 regression tress when the ensemble mean is taken
by the use of Algorithm \ref{Alg: DTCV} versus simple uniform averaging.
These are called DT and Uniform, respectively, in the table. The values
in parenthesis correspond to $\mathcal{T}=\mathcal{U}$. The bold
fonts show better losses in each category; weak vs strong.}

\label{Tab: Overall}
\end{table}

\begin{table}[t]
\begin{tabular}{|c|c|c|c|c|}
\hline 
Max Fold & \multicolumn{1}{c}{Weak Trees} &  & \multicolumn{1}{c}{Strong Trees} & \tabularnewline
\hline 
 & Uniform & DT & Uniform & DT\tabularnewline
\hline 
Housing & 2.91(2.93) & \textbf{2.40}(\textbf{2.52}) & 1.74(1.72) & \textbf{1.73}(\textbf{1.70})\tabularnewline
\hline 
Grid & 32.25(31.96) & \textbf{29.97}(\textbf{27.47}) & 9.79(\textbf{8.50}) & \textbf{9.74}(\textbf{8.50})\tabularnewline
\hline 
SC & 21.96(19.22) & \textbf{15.61}(\textbf{12.56}) & 5.40(5.23) & \textbf{5.39}(\textbf{5.12})\tabularnewline
\hline 
Bike & 19.43(17.60) & \textbf{11.98}(\textbf{10.10}) & \textbf{1.31}(1.38) & 1.34(\textbf{1.36})\tabularnewline
\hline 
\end{tabular}

\caption{\textbf{Maximum Fold Losses Comparison}. This table shows the maximum
fold loss comparison of 10 regression trees. DT uses  the distribution over models  obtained from
 Algorithm \ref{Alg: DTCV}. Uniform refers to the uniform distribution over models. These are called DT and Uniform, respectively, in the table.
The values in parenthesis correspond to $\mathcal{T}=\mathcal{U}$.
The bold fonts show better losses in each category; weak vs. strong.}

\label{Tab: MaxFold}
\end{table}

\begin{figure}[t]
\includegraphics[scale=0.5]{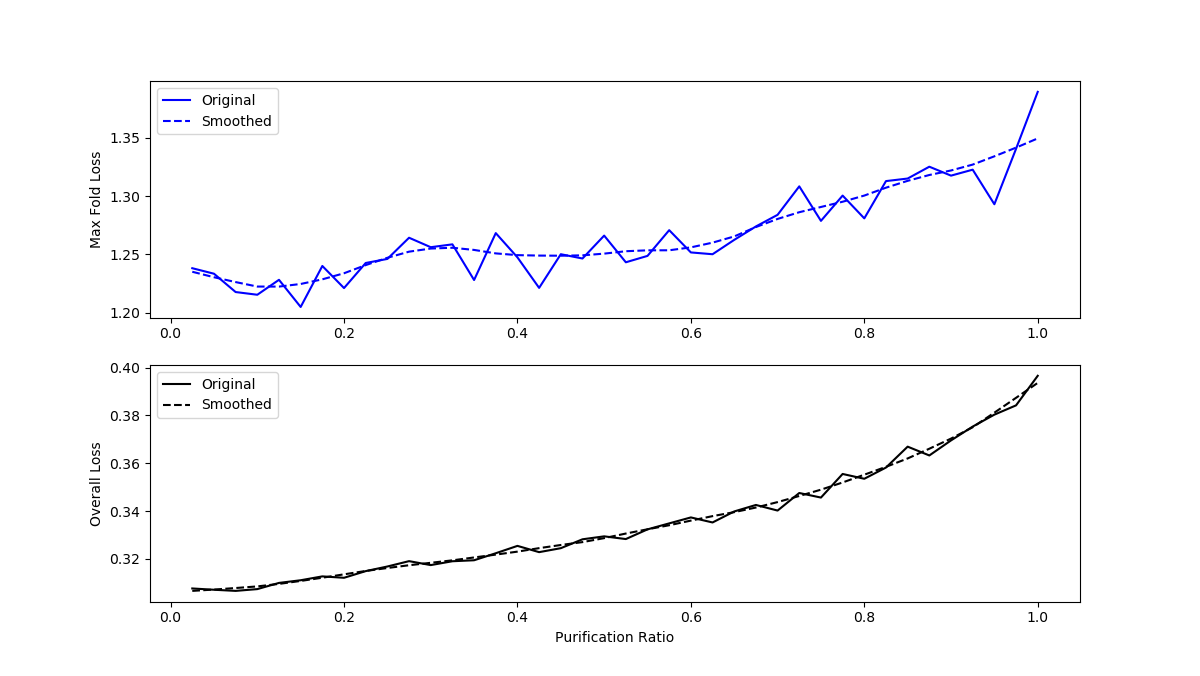}

\caption{\textbf{Purification Ratio}. The \emph{top plot} shows the maximum
fold loss for different purification ratios. The \emph{bottom plot}
shows the overall loss for different purification ratios. Both plots
correspond to the test dataset. A weighted local linear regression
is used to provide a better picture in both plots. These plots show
that the purification ratio should be around $0.15$ for this
particular dataset to achieve the best balance between the overall loss and maximum fold loss.}

\label{Fig: Purification Ratio}
\end{figure}

\begin{example}
In this example, we investigate the effect of $ns/\left|\mathcal{U}\right|$.
Since this ratio is a part of the purification over repeated games
\cite{harsanyi1973games}, we call it ``purification ratio'' for
brevity. Our purpose is to illustrate the maximum fold loss, and also
the overall loss, as a function of the purification
whose values are between $0$ and $1$. We work with the California Housing
\cite{pace1997sparse} dataset, and we take an ensemble of models produced
by Algorithm \ref{Alg: DTCV} with $n=100$ . In all the cases, we
take $10$ regression trees with a maximum depth of $10$. The original
data is divided into $80\%$ train-UQ and $20\%$ test sets. Each
tree is trained on randomly selected $50\%$ of the train set. This
time, to estimate the robustness of the algorithm using maximum fold
loss, we use only $20$ bins (folds) over the test set. For each purification
ratio, we conduct the experiment $100$ times, and we plot the mean
of each experiment. To make sure enough data points are seen by the estimators,
$K$ is taken to be $5$ over the purification ratio.

The results are shown in Figure \ref{Fig: Purification Ratio}. As can be seen from the plots, 
there is a purification ratio close to $0.15$ that produces the minimal maximum fold loss in all the tests. At the same time, the overall loss decreases uniformly with a decrease in the purification ratio. This experiment shows that to capture the best balance between 
the overall loss and maximum fold loss, the purification ratio $ns/\left|\mathcal{U}\right|$ 
should be around $0.15$. This also supports our choice in the previous example of 0.20.
\end{example}

\begin{figure}[t]
\includegraphics[scale=0.5]{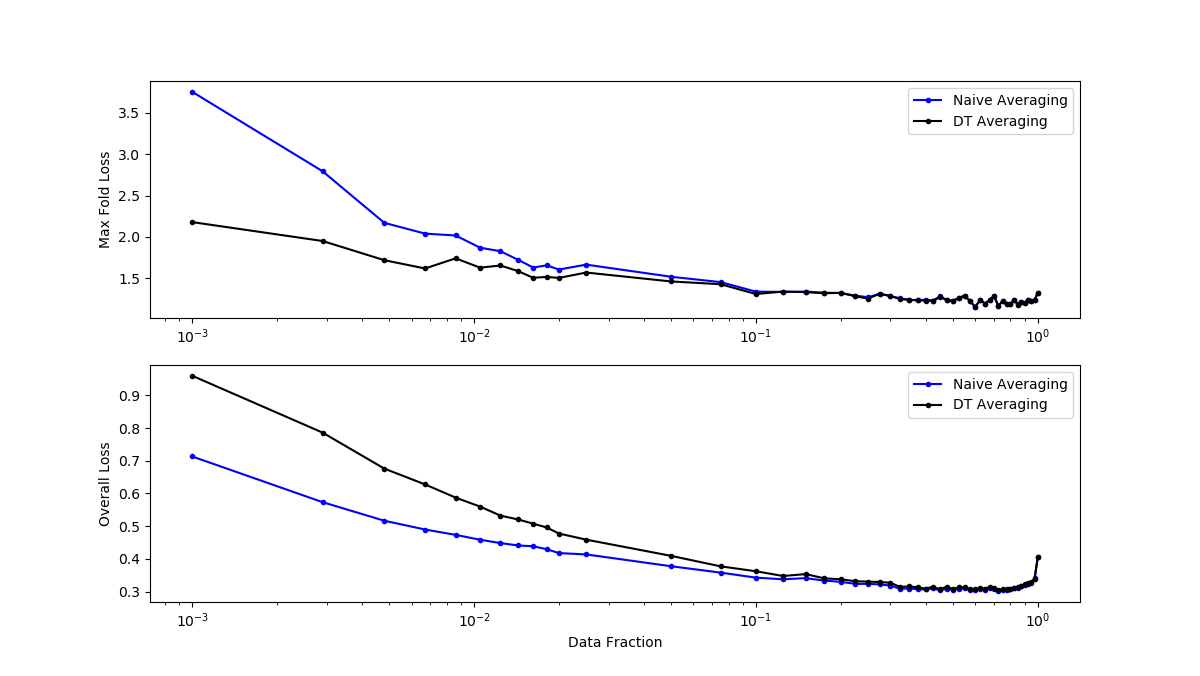}

\caption{\textbf{Small vs Large Data Test}. The \emph{top plot} shows that
Algorithm \ref{Alg: DTCV} is robust when the available data is sparse.
The \emph{bottom plot} shows that uniform averaging always provide a
more accurate prediction. Both plots show that for large datasets,
both approaches converge in robustness and accuracy.}

\label{Fig: Data Fraction}
\end{figure}

\begin{example}
In this example, we investigate the effect of data sparsity on Algorithm
\ref{Alg: DTCV}. As in the previous
example, we work with the California Housing \cite{pace1997sparse}
dataset. In all the cases, we take $10$ regression trees with a maximum
depth of $10$. The original data is divided into $80\%$ train-UQ
and $20\%$ test sets. The train-UQ set is then fed into Algorithm
\ref{Alg: DTCV} with $\mathcal{T}=\mathcal{U}$. The latter is done
for the sake of a fair comparison with uniform ensembles. We take $n$
and $K$ to be both $100$. Also, the purification ratio is taken
to be $0.20$. The maximum fold loss is calculated over $20$ bins
of the test set. In this experiment, we vary the data availability
by allowing a fixed fraction of the train set to each estimator. We
call this variable the ``data fraction''. Figure \ref{Fig: Data Fraction}
shows that for sparse data availability, Algorithm \ref{Alg: DTCV}'s error
is unanimously less than the uniform averaging method. At the same time,
the overall loss is always smaller for uniform averaging. This figure
conveys the message that Algorithm \ref{Alg: DTCV} is more robust
than uniform averaging, specifically when only a small amount of data
is available.
\end{example}

\section{Related Works\label{sec:Related-Works}}

\subsection{Sources of Uncertainty: \label{subsec:Sources-of-Uncertainty:}}

From a statistical perspective, uncertainty manifests itself in two
general forms. 

\subsubsection{Aleatoric Uncertainty (Irreducible Uncertainty):}

This type of uncertainty is intrinsic to the randomness of the data
generating process and is not reducible \cite{der2009aleatory}. It
is also known as the \emph{statistical uncertainty}. Aleatoric uncertainty
can be categorized into homoskedastic and heteroskedastic categories
\cite{diebold1998elements}. The homoskedastic randomness stays constant
for all data points, however, the heteroskedastic uncertainty is variant
with respect to them. 

\subsubsection{Epistemic Uncertainty (Reducible Uncertainty):\label{subsec:Epistemic-Uncertainty-(Reducible}}

This type of error is about the oversight on data and can be reduced
with an increase in data collection, incorporation of more features,
and a better selection of the model \cite{der2009aleatory}. 

\subsection{What does UQ mean in ML?\label{subsec:What-does-UQ}}

Based on the sources of uncertainty enumerated, we can define UQ as
any method that can quantify the individual or combined effects of
the aleatoric and epistemic uncertainties. Hence, UQ can have different
interpretations in the context of ML, and statistical learning theory
\cite{friedman2001elements}. In what follows, we express a survey
of some of these methods.

\subsubsection{UQ as Expected Generalization (Prediction) Error:\label{subsec:UQ-as-Expected}}

In this form, UQ quantifies the expected generalization error of a
machine learning algorithm. There are two separate approaches here.
One is analytical, used for linear models, and based on the training
dataset (in-sample) \cite{montgomery2015introduction}. The other
is based on efficient sample re-use (extra-sample) and can be used
for both linear and nonlinear models \cite{friedman2001elements}.
The first approach encompasses methods like Mallows' $C_{p}$ \cite{mallows1995more,mallows1973some,mallows1964choosing,mallows1967choosing},
and Akaike information criterion (AIC) \cite{akaike1998information}.
The second approach uses techniques like cross-validation (CV) \cite{stone1974cross,stone1977asymptotic,allen1974relationship,golub1979generalized,wahba1980spline}
and  bootstrapping \cite{efron1983estimating,breiman1996bagging,breiman2001random,quinlan2014c4}. 
For small to moderately large models, CV methods can be used readily. 
However, computational costs may impede such approaches for larger models like deep neural networks (DNN).

\subsubsection{UQ as Posterior Distribution Variance:\label{subsec:UQ-as-Posterior}}

In this form, the machine learning that is used must be probabilistic
in a Bayesian inference (BI) framework. The posterior distribution
of the model parameters can then be exploited to express prediction
confidence intervals \cite{gelman2013bayesian,murphy2012machine}.
The major issue here is that the Bayesian approach needs priors that
are not necessarily found from observations. For linear models, this
type of UQ task is commonly done by the Bayesian information criterion
(BIC) \cite{schwarz1978estimating}. For nonlinear models, it is done
by approaches like the Monte Carlo (MC) and Markov Chain Monte Carlo
(MCMC) \cite{geman1984stochastic,gelfand1990sampling,gilks1995markov,gelman2013bayesian}.
Unfortunately, the MCMC methods become computationally expensive for
larger models. Using a variational Bayesian
inference (VBI) \cite{graves2011practical} has been proposed as a
possible solution. For example Bayesian neural networks (BNN) \cite{mackay1992practical,neal2012bayesian,graves2011practical,louizos2017multiplicative,blundell2015weight,rudin2019stop}
are mainly using VBI. However, the computational costs are still prohibitive
\cite{ovadia2019can}, the MCMC comparisons show the low quality of posterior
approximations \cite{yao2019quality}, and these approaches are sensitive
to initial starting values \cite{rossi2018good}.

\subsubsection{UQ as Regression Quantiles:\label{subsec:UQ-as-Regression}}

There are methods in which data is used to provide quantile bounds
as prediction intervals over regressors in a frequentist sense \cite{koenker2001quantile,koenker1978regression}.
An extension of this methodology is the method of quantile regression
forests (QRF) estimating the predictions of a regressor in terms of
quantile values \cite{meinshausen2006quantile}. Recently, conformal
quantile regression (CQR) has combined conformal prediction \cite{papadopoulos2002inductive,vovk2005algorithmic,papadopoulos2008inductive}
with quantile regression (QR) methodology \cite{meinshausen2006quantile,koenker2001quantile,koenker1978regression}
to provide confidence intervals by using a ``pinball loss'' \cite{koenker1978regression}
over a train set and then applying a conformal prediction over a calibration
set \cite{romano2019conformalized}. Although this method addresses
heteroskedastic data with outliers, it requires a specific loss function. 

\subsubsection{Deep Learning UQ (DLUQ ) Survey:\label{subsec:Deep-Learning-UQ}}

Deep learning (DL) has arisen as a particularly powerful technology
due to its ability to ingest raw data with less preprocessing than
classical ML \cite{goodfellow2016deep,lecun2015deep}. However, DL
models are costly at the training phase. Hence, the application of
the UQ views expressed in this section is not straightforward. Because
of this major issue, the deep neural networks (DNN) models become over-confident
\cite{nguyen2015deep}. Ad hoc methods like confidence calibration
\cite{guo2017calibration}, temperature scaling \cite{hinton2015distilling}
( using Platt scaling \cite{platt1999probabilistic,niculescu2005predicting}),
histogram binning \cite{zadrozny2001obtaining,zadrozny2002transforming},
isotonic regression \cite{zadrozny2002transforming}, deep ensembles
with Gaussian mixture \cite{lakshminarayanan2017simple}, and Bayesian
binning into quantiles \cite{naeini2015obtaining} are introduced
to address deep learning UQ. Unfortunately, all these methods are
mainly post-processing steps to models. The only systematic DLUQ is
the dropout UQ approach \cite{gal2016dropout,srivastava2014dropout}.
Although this method is closely related to BI of deep Gaussian processes
\cite{rasmussen2006gaussian,damianou2013deep}, it is a limit case
study that could deviate from a Gaussian process for a finite width-depth
DNN. 

\section{Conclusion\label{sec:Discussion-and-Future}}

In this article, we have furnished a systematic machine learning UQ
approach based on decision theory. DTB provides a rigorous
UQ in a cross-validation setting and produces ML algorithms that are
more stable with respect to distribution shifts. The algorithm presented
in this paper can show its best performance when there is significant
epistemic uncertainty. At the same time, DTB provides pointwise
prediction confidence intervals for ML regressors. This framework
is not confined to regression problems in a classical setting and
can be readily extended to DNN and classification tasks as well. DTB
slightly sacrifices accuracy to achieve high stability for
ML estimators. 

\section*{Acknowledgment\label{sec:Acknowledgment}}

This research was carried out at the Jet Propulsion Laboratory, California
Institute of Technology, under a contract with the National Aeronautics
and Space Administration and support from Beyond Limits (Learning Optimal Models) and AFOSR (Grant number FA9550-18-1-0271, Games for Computation and Learning). The authors are thankful to Amy Braverman,
Lukas Mandrake and Kiri Wagstaff, for their insights.

\textsuperscript{\textcopyright} 2021. California Institute of Technology. Government sponsorship
acknowledged. 

\bibliographystyle{plain}
\bibliography{DTCS}

\end{document}